\documentclass[sigconf]{acmart}
\usepackage{balance} 
\usepackage{enumitem}
\usepackage{titlecaps}
\usepackage{mfirstuc}
\usepackage{bm}
\usepackage{enumitem}
\usepackage{wrapfig}
\usepackage{adjustbox}
\usepackage{amsmath}
\usepackage{multicol}
\usepackage{multirow}
\usepackage{subcaption}

\usepackage{amssymb}
\usepackage{amssymb}
\usepackage{booktabs} 
\usepackage{xcolor,colortbl}
\usepackage{float}
\hypersetup{
    colorlinks=true,     
    citecolor=blue,     
    linkcolor=red,       
    urlcolor=blue,      
    filecolor=purple    
}
\AtBeginDocument{%
  }
\copyrightyear{2025}
\acmYear{2025}
\setcopyright{acmlicensed}\acmConference[ICMR '25]{Proceedings of the 2025
International Conference on Multimedia Retrieval}{June 30-July 3,
2025}{Chicago, IL, USA}
\acmBooktitle{Proceedings of the 2025 International Conference on Multimedia
Retrieval (ICMR '25), June 30-July 3, 2025, Chicago, IL, USA}
\acmDOI{10.1145/3731715.3733478}
\acmISBN{979-8-4007-1877-9/2025/06}

\begin{document}
\definecolor{yellow}{rgb}{1, 1, 0.7}
\definecolor{orange}{rgb}{1, 0.85, 0.7}
\definecolor{red1}{rgb}{1, 0.7, 0.7}
\title{GarmentGS: Point-Cloud Guided Gaussian Splatting for High-Fidelity Non-Watertight 3D Garment Reconstruction}

\author{Zhihao Tang}
\orcid{0009-0007-4779-0088}
\affiliation{%
\institution{ Donghua University}
  \state{Shanghai}
  \country{China}}
\email{220995117@mail.dhu.edu.cn}

\author{Shenghao Yang}
\orcid{0009-0006-3943-7914}
\affiliation{%
\institution{ Donghua University}
  \state{Shanghai}
  \country{China}}
\email{2242164@mail.dhu.edu.cn}

\author{Hongtao Zhang}
\orcid{0000-0001-9868-1243}
\authornote{Corresponding authors.}
\affiliation{
\institution{Donghua University}
  \state{Shanghai}
  \country{China}}
\email{2201957@mail.dhu.edu.cn}

\author{Mingbo Zhao}
\orcid{0000-0003-0381-4360}
\authornotemark[1]
\affiliation{%
\institution{ Donghua University}
  \state{Shanghai}
  \country{China}}
\email{mzhao4@dhu.edu.cn}




\renewcommand{\shortauthors}{Zhihao Tang, Shenghao Yang, Hongtao Zhang, and Mingbo Zhao}


\begin{abstract}
Traditional 3D garment creation requires extensive manual operations, resulting in time and labor costs. Recently, 3D Gaussian Splatting has achieved breakthrough progress in 3D scene reconstruction and rendering, attracting widespread attention and opening new pathways for 3D garment reconstruction. However, due to the unstructured and irregular nature of Gaussian primitives, it is difficult to reconstruct high-fidelity, non-watertight 3D garments. In this paper, we present GarmentGS, a dense point cloud-guided method that can reconstruct high-fidelity garment surfaces with high geometric accuracy and generate non-watertight, single-layer meshes. Our method introduces a fast dense point cloud reconstruction module that can complete garment point cloud reconstruction in 10 minutes, compared to traditional methods that require several hours. Furthermore, we use dense point clouds to guide the movement, flattening, and rotation of Gaussian primitives, enabling better distribution on the garment surface to achieve superior rendering effects and geometric accuracy. Through numerical and visual comparisons, our method achieves fast training and real-time rendering while maintaining competitive quality. 
\end{abstract}




\maketitle

\begin{figure}[htb]
\centering
\captionsetup{skip=2mm}
\includegraphics[width=\linewidth]{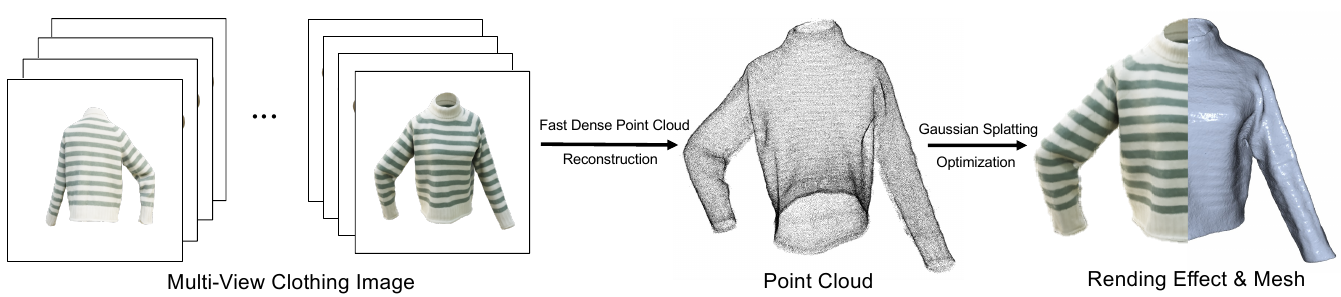}
    \caption{Our method reconstructs 3D garments from multi-view images through fast dense point cloud reconstruction and Gaussian primitive optimization, achieving high-fidelity rendering effect while generating non-watertight meshes that preserve complex topological.} 
    \label{page1}
\vspace{-10pt}
\end{figure}

\section{Introduction}

With the increasing popularity of applications such as fashion design, virtual wardrobes, and virtual try-on, the importance of 3D garment assets has become increasingly prominent. However, traditional workflows remain labor-intensive and inefficient, relying on
skilled professionals and specialized software ~\cite{MAYA, CLO3D, Style3D} to manually create models over weeks. 
Current 3D garment generation methods face significant technical trade-offs. Traditional approaches, including data-driven generative models~\cite{get3d,Shap-e} and 2D sewing pattern-based methods~\cite{he2024dresscode,liu2023towards}, struggle with limited fabric detail, stylistic uniformity, and reliance on extensive paired training data. While NeRF-based methods~\cite{nerf} improve reconstruction quality through neural rendering, their slow computational speed and implicit representations hinder industrial adoption. 

Recent advances in 3D Gaussian Splatting~\cite{kerbl3DGaussianSplatting2023} address some of these issues by offering explicit Gaussian-based representations, enabling faster training, real-time rendering, and easier mesh conversion for pipeline integration. However, critical challenges remain unresolved: existing methods like 2DGS~\cite{huang2DGaussianSplatting2024} and PGSR~\cite{chen2024pgsrplanarbasedgaussiansplatting} cannot create wearable garments with functional openings (e.g., necklines, sleeves), and reconstructed meshes often exhibit fragmented internal layers or multi-layered artifacts. These issues significantly impact the final quality of 3D garments and make the reconstructed garments difficult to use directly in subsequent industrial pipelines.

To address these challenges, we present \textbf{GarmentGS}, an innovative method based on 3DGS~\cite{kerbl3DGaussianSplatting2023} that generates high-fidelity, non-watertight, single-layer 3D garments from multi-view clothing images that can be directly applied to subsequent workflows. First, we developed a fast MVS-based dense point cloud reconstruction method. Compared to traditional reconstruction methods~\cite{yang2003multi,xu2019multi,galliani2015massively,schonberger2016structure} that typically take several hours, our method requires only about 10 minutes to generate dense point clouds from multi-view garment images. Next, inspired by GaussianPro~\cite{cheng2024gaussianpro}, we utilize the generated dense point clouds to adjust the position and rotation of Gaussian ellipsoids, enabling better distribution on the garment surface, thereby enhancing geometric reconstruction and achieving non-watertight reconstruction. Inspired by 2DGS~\cite{huang2DGaussianSplatting2024}, we flatten 3D Gaussian ellipsoids into 2D Gaussian elliptical disks to further improve geometric reconstruction quality. Finally, after mesh extraction, we apply LOF~\cite{breunig2000lof} in conjunction with dense point clouds to optimize the mesh, removing internal fragmented faces to obtain a single-layer mesh suitable for subsequent fabric simulation. In summary, our main contributions are as follows:
\begin{itemize}[leftmargin=*]
\item We propose a fast MVS-based dense point cloud reconstruction method that efficiently generates dense point clouds from multi-view garment images.
\item We propose a new Gaussian primitives optimization method by utilizing the generated dense point clouds, achieving non-watertight reconstruction and improving overall reconstruction quality.

\item We propose a method that applies dense point clouds to trim and denoise the reconstructed mesh, resulting in a single-layer mesh suitable for fabric simulation.
\end{itemize}

\section{Related Work}
\label{sec:format}
\subsection{Multi-View Stereo}
\label{ssec:subhead}
Multi-view stereo (MVS) algorithms create 3D models by integrating views from different viewpoints, enabling accurate reconstruction in complex environments. With its efficiency and reliability, MVS has become the cornerstone method for image-based 3D reconstruction.

Recently, deep learning has improved MVS methods~\cite{wang2018mvdepthnet,yao2018mvsnet,kim2021just}, with depth map-based approaches being the most popular. These methods~\cite{wang2018mvdepthnet,yao2018mvsnet} estimate depth maps for individual images through patch matching, then fuse them into dense representations like point clouds or meshes. This approach improves flexibility and scalability by separating depth estimation and fusion. However, it requires significant computational resources due to dense pixel matching and multi-view consistency optimization. 


To accelerate dense point cloud reconstruction, we adopted a simplification strategy: removing color information and processing only geometric information, while also reducing depth map resolution and decreasing the number of multi-view consistency optimization iterations, thereby  reducing computational overhead.
\subsection{Surface Reconstruction with 3D Gaussians}
\label{ssec:subhead}
3D Gaussian Splatting~\cite{kerbl3DGaussianSplatting2023}offers fast rendering, intuitive representation, and high-quality output, making it a new paradigm in neural rendering. However, due to chaotic, noisy, and uneven distribution of 3D Gaussian ellipsoids, it achieves high-fidelity rendering but often sacrifices geometric accuracy. 

To address this issue, researchers have proposed two main approaches: one converts 3D Gaussian ellipsoids into 2D Gaussian elliptical disks through a series of regularization terms~\cite{huang2DGaussianSplatting2024,chen2024pgsrplanarbasedgaussiansplatting}, ensuring alignment of Gaussian primitives with object surfaces - while sacrificing rendering quality, it enhances geometric reconstruction effects; the other combines 3D Gaussians with neural signed distance fields (SDF)~\cite{yuGSDF3DGSMeets2024,zhang2024neural}, building a dual-branch architecture that enhances reconstruction and rendering effects through mutual guidance and joint supervision during training.

Although these methods have achieved good results in surface geometric reconstruction, limitations still exist in garment reconstruction. While regularization-based methods~\cite{huang2DGaussianSplatting2024,chen2024pgsrplanarbasedgaussiansplatting} can achieve high-fidelity garment surface reconstruction, they cannot achieve non-watertight garment reconstruction. And while SDF-combined methods~\cite{yuGSDF3DGSMeets2024,zhang2024neural} can accurately position Gaussian primitives on garment surfaces, partially achieving non-watertight reconstruction, the training speed issues of SDF result in the loss of 3D Gaussian's fast rendering advantage. Our method chooses to combine 3DGS with dense point cloud priors rather than SDF, achieving both excellent surface geometric reconstruction and non-watertight garment reconstruction while maintaining fast rendering.

\begin{figure*}
\centering
\includegraphics[width=\linewidth]{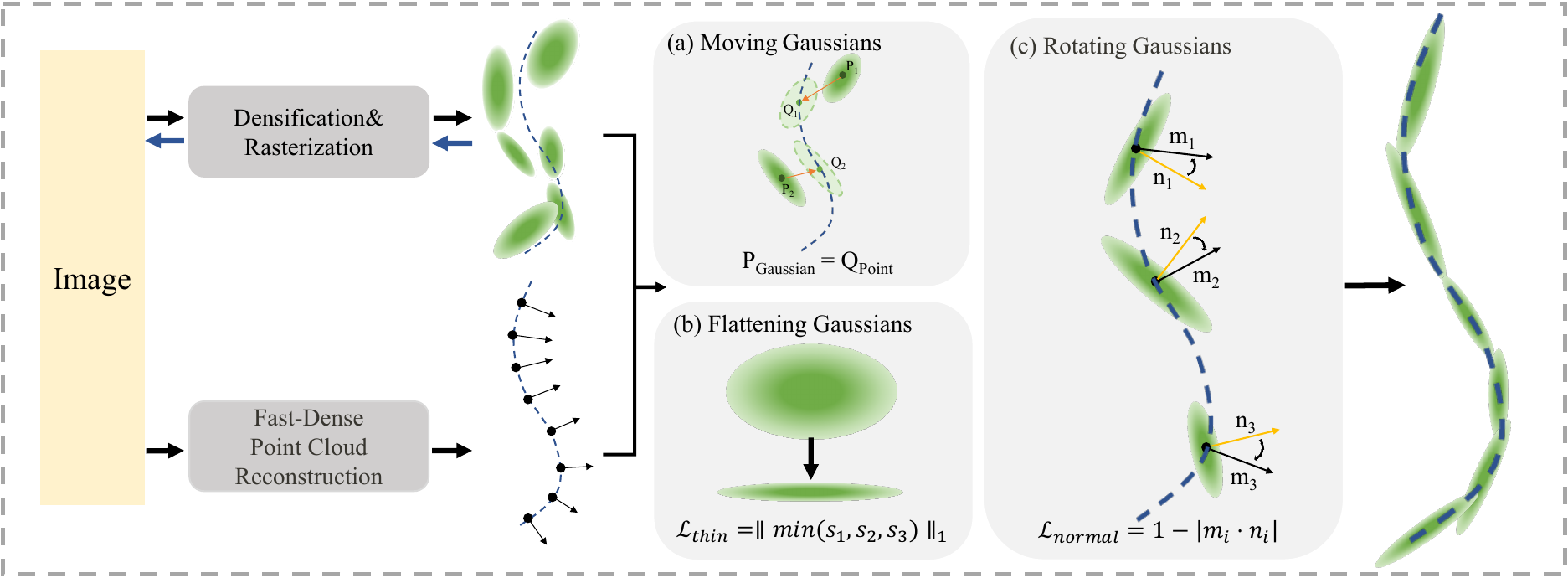}
    \caption{Overview of our method. We first \textbf{(a)} pull scattered 3D Gaussian primitives towards their nearest points in the dense point cloud, while \textbf{(b)} flattening 3D Gaussian ellipsoids into 2D Gaussian elliptical disks, then \textbf{(c)} rotate the 2D Gaussian elliptical disks at corresponding positions according to the normal directions of points in the dense point cloud to align the Gaussian primitives precisely with the clothing surface.} 
    \label{pipline}
\vspace{-10pt}
\end{figure*} 
\section{Method}
\label{sec:pagestyle}

We propose a Gaussian optimization framework based on dense point clouds, which optimizes Gaussian primitives through dense point clouds, as shown in Fig.~\ref{pipline}. This method uses dense point clouds to precisely align Gaussian primitives with garment surfaces and flattens 3D Gaussian ellipsoids into 2D Gaussian elliptical disks, thereby improving geometric reconstruction quality but also achieving non-watertight reconstruction. Additionally, we utilize dense point clouds to precisely trim the reconstructed mesh, removing internal fragment faces to obtain a single-layer mesh.

\subsection{Preliminary}
\label{ssec:subhead}
\textbf{3D Gaussian Splatting.} 3DGS~\cite{kerbl3DGaussianSplatting2023} has become an important method for 3D representation with its high-fidelity rendering quality and real-time rendering capabilities. It defines points in point clouds as 3D Gaussian primitives with volumetric density:
\begin{equation}
    G(\bm{x})=\exp{\left( -\frac{1}{2} (\bm{x})^T \bm{\Sigma^{-1}} (\bm{x}) \right)},
\end{equation}
where $\bm{\Sigma}$ is the 3D covariance matrix and $\bm{x}$ is the position relative to point $\bm{\mu}$.
To ensure the positive semi-definiteness of the covariance matrix, 3DGS transforms it into a combination of rotation matrix $\bm{R_i}$ and scaling matrix $\bm{S_i}$:
\begin{equation}
    \bm{\Sigma_i} = \bm{R_i} \bm{S_i} \bm{S_i}^T \bm{R_i}^T.
\end{equation}
In addition to geometric properties, each Gaussian primitive also stores opacity $\bm{\alpha}$ and a set of learnable spherical harmonics (SH) parameters to represent view-dependent appearance features. Furthermore, the gradient-based Gaussian adaptive control mechanism plays a key role in improving scene representation accuracy, reducing redundancy, and optimizing Gaussian primitives to better match the underlying geometry and appearance.

\textbf{Dense Point Cloud Reconstruction.} Dense point cloud reconstruction based on depth maps requires pixel-level matching and multi-view consistency optimization, which incurs significant computational overhead when processing multiple images. To improve reconstruction efficiency, we simplified the algorithm pipeline: retaining only geometric information processing (position and normals), reducing depth map resolution, and decreasing optimization iteration counts. Experiments show that when processing the same 100 multi-view images, our method completes reconstruction in just 10 minutes, while traditional method~\cite{schonberger2016pixelwise} require 2 hours.

\subsection{Gaussian Optimization}
\label{ssec:subhead}
\textbf{Moving Gaussians.} To achieve better geometric reconstruction effects and non-watertight reconstruction, we need to position 3D Gaussian primitives on the garment surface. To this end, as shown in Fig.~\ref{pipline}(\textbf{a}), we utilize dense point clouds as priors to guide the movement of Gaussian primitives. Specifically, we first construct a K-D tree~\cite{bentley1975multidimensional} using dense point clouds, then apply the Euclidean distance formula for each Gaussian primitive:
\begin{equation}
    d(\bm{P}, \bm{Q}) = \sqrt{\sum_{i=1}^{3} (p_i - q_i)^2}
\end{equation}
to query the K-D tree to find the nearest point in the point cloud, where $\bm{P}$ is the mean value of the 3D Gaussian and $\bm{Q}$ is the coordinate of the point in the point cloud. Afterwards, we update the position of the Gaussian primitive:
\begin{equation}
    \bm P_{Gaussian} = \bm Q_{Point},
\end{equation}
where $\bm Q_{Point}$ is the point nearest to $\bm P_{Gaussian}$.

\textbf{Flattening Gaussians.} The covariance matrix $\bm{\Sigma_i} = \bm{R_i} \bm{S_i} \bm{S_i}^T \bm{R_i}^T$ of 3DGS represents the shape of the Gaussian ellipsoid, where $\bm{R_i}$ represents the orthogonal basis of the ellipsoid's three axes, and $\bm{S_i}$ represents the scale along each direction. According to the method, by compressing $\bm{S_i}$ along a specific axis, the Gaussian ellipsoid can be flattened into a plane aligned with that axis. When compression is performed along the direction of smallest $\bm{S_i}$, the Gaussian ellipsoid is compressed into a plane that most closely matches its original shape. Therefore, as shown in Fig.~\ref{pipline}(\textbf{b}), we minimize $\bm{S_i}$ = $\bm{diag(s_1,s_2,s_3)}$ for each Gaussian ellipsoid:
\begin{equation}
    \L_{thin} = \parallel \min(s_1, s_2, s_3) \parallel_1.
\end{equation}

\textbf{Rotating Gaussians.} To ensure accurate surface fitting of the Gaussian elliptical disks and improve geometric reconstruction accuracy, we align the normals of the Gaussian elliptical disks with the normals of points in the dense point cloud. As shown in Fig.~\ref{pipline}(\textbf{c}), we align the normal $\bm{n_i}$ of Gaussian elliptical disk $\bm{P_i}$ with the normal $\bm{m_i}$ of its corresponding point $\bm{Q_i}$:
\begin{equation}
    \L_{normal} = 1-|m_i\cdot n_i|.
\end{equation}

\textbf{Final Loss.} We minimize the following loss function:
\begin{equation}
    \L = \L_{RGB} + \alpha \L_{thin} + \beta \L_{normal} ,
\end{equation}
where $\L_{RGB}$ is the RGB reconstruction loss combining $\L_{1}$ with the D-SSIM term in 3DGS~\cite{kerbl3DGaussianSplatting2023}, $\L_{thin}$ is the loss for Gaussian flattening, and $\L_{normal}$ is the loss for Gaussian rotating. We set $\bm{\alpha}$ to 100 and $\bm{\beta}$ to 0.1.

\subsection{Mesh Denoising}
\label{ssec:subhead}
After completing mesh extraction, we found that although non-watertight garment reconstruction was achieved, the extracted mesh exhibited an inner-outer double-layer structure, with the inner mesh primarily composed of fragmented faces. To address this issue, we use LOF~\cite{breunig2000lof} to detect and remove fragmented faces. Since points in the dense point cloud are strictly distributed on the surface, they are highly suitable for training the LOF model. We traverse each vertex of the mesh using the trained LOF model, remove faces corresponding to outlier points, and ultimately obtain a single-layer mesh suitable for fabric simulation.

\begin{figure}[htb]
\centering
\includegraphics[width=\linewidth]{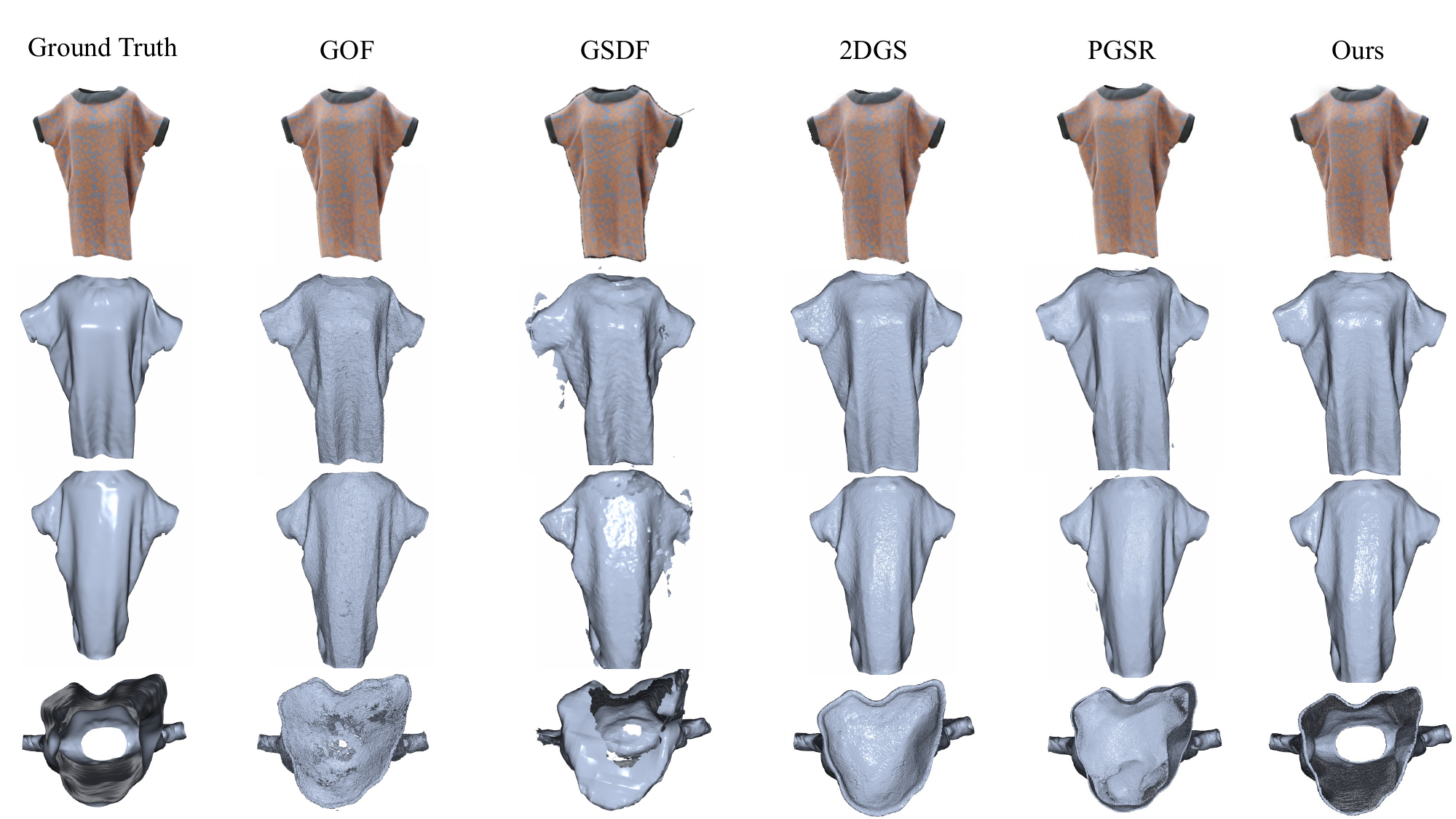}
    \caption{Comparison between ours and other methods on multi-view reconstruction on DeepFashion3D dataset. Top row: rendering effect. Bottom 3 rows: reconstructed meshes.} 
    \label{visual}
\vspace{-10pt} 
\end{figure}
\begin{table}
  \centering
  \caption{Rendering and reconstruction comparisons against baselines. Best results are highlighted as \colorbox{red1}{1st}, \colorbox{yellow}{2nd} and \colorbox{orange}{3rd}.}
  \label{tab:table1}
  \renewcommand{\arraystretch}{1.2}
  \setlength{\tabcolsep}{9pt}
\resizebox{\linewidth}{!}{ 
  \begin{tabular}{l|cccccccc}
    \toprule
      & SSIM$\uparrow$ & PSNR$\uparrow$ & LPIPS$\downarrow$ & CD$\downarrow$ & Time$\downarrow$ & Single-layer & Wearable & Texture   \\
    \midrule
     3DGS~\cite{kerbl3DGaussianSplatting2023} & \ 0.842 & 23.61 & 0.057 & -- & \cellcolor{red1}3.3m &--  &--  &--  \\  
     GOF~\cite{yu2024gaussianopacityfieldsefficient} & \cellcolor{red1}0.966 & 36.24 & \cellcolor{yellow}0.026 & \cellcolor{orange}0.735 & 22.5m &--  & -- & $\checkmark$ \\
     GSDF~\cite{yuGSDF3DGSMeets2024} & \cellcolor{yellow}0.965 & \cellcolor{yellow}37.14 & 0.028 & 1.034 & $\sim$~1h &--  &--  & $\checkmark$  \\
     2DGS~\cite{huang2DGaussianSplatting2024} & \cellcolor{yellow}0.965 & \cellcolor{orange}36.33 & \cellcolor{orange}0.027 &\cellcolor{yellow}0.712 & \cellcolor{yellow}3.9m &--  &--  & $\checkmark$  \\    
     PGSR~\cite{chen2024pgsrplanarbasedgaussiansplatting}& \cellcolor{orange}0.961 & 35.22 & 0.031 & 0.737 & 13.9m  &--  &--  & $\checkmark$  \\
     Ours & \cellcolor{yellow}0.965 & \cellcolor{red1}40.13 & \cellcolor{red1}0.017 & \cellcolor{red1}0.564 & \cellcolor{orange}11.3m & $\checkmark$ & $\checkmark$ & $\checkmark$  \\
  \bottomrule
    \end{tabular}}
\vspace{-10pt}  
\end{table}

\section{Experiments}
\label{sec:majhead}
\subsection{Experiment Settings}
\label{ssec:subhead}
\textbf{Datasets.} We use the DeepFashion3D-v2~\cite{zhuDeepFashion3DDataset2020} to quantitatively evaluate GarmentGS's rendering and reconstruction performance. Consistent with previous method~\cite{liu2023ghost}, we select 9 ground truth mesh models from DeepFashion3D-v2, and use Blender's Cycles engine with real environment lighting maps to render 100 views for training and testing data.

\textbf{Evaluation Metrics.} We use Chamfer Distance (CD) to evaluate the accuracy of reconstructed meshes. To evaluate rendering quality, we employ PSNR, SSIM, and LPIPS metrics. Additionally, we assess three key characteristics of the reconstructed meshes: single-layer structure, wearability (non-watertight), and the presence of texture mapping.

\textbf{Baselines.} We compare our method with the latest GS-based reconstruction methods, including 3DGS~\cite{kerbl3DGaussianSplatting2023}, GOF~\cite{yu2024gaussianopacityfieldsefficient}, GSDF~\cite{yuGSDF3DGSMeets2024}, 2DGS~\cite{huang2DGaussianSplatting2024}, and PGSR~\cite{chen2024pgsrplanarbasedgaussiansplatting}.

\textbf{Implementation Details.} Our training strategy and hyperparameters generally follow the settings of 3DGS. All scenes are trained for 30,000 iterations. All experiments in this paper were conducted on an Nvidia RTX 3090 GPU.

\begin{figure}[htb]
\centering
\includegraphics[width=0.9\linewidth]{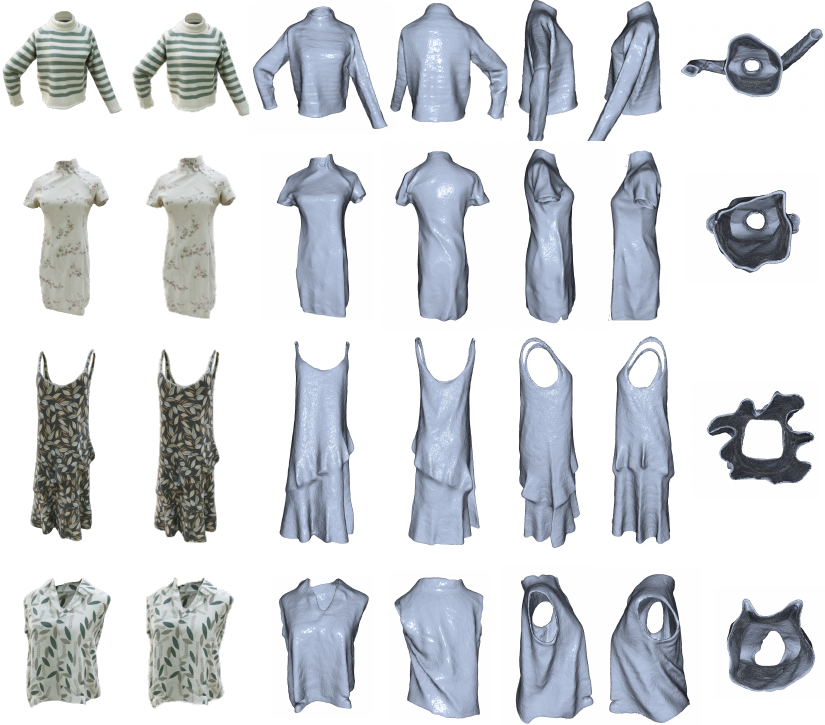}
    \caption{Results for some garments. First column: ground truth. Second column: rendering effect. Remaining 5 columns: reconstructed meshes.} 
    \label{visual1}
\vspace{-10pt}  
\end{figure}

\subsection{Comparisons}
\label{ssec:subhead}
In Table \ref{tab:table1}, we present a detailed comparison of the performance between 3DGS~\cite{kerbl3DGaussianSplatting2023}, GOF~\cite{yu2024gaussianopacityfieldsefficient}, GSDF~\cite{yuGSDF3DGSMeets2024}, 2DGS~\cite{huang2DGaussianSplatting2024}, PGSR~\cite{chen2024pgsrplanarbasedgaussiansplatting}, and our method on the DeepFashion3Dv2~\cite{zhuDeepFashion3DDataset2020} dataset, including various metrics for garment rendering (SSIM, PSNR, LPIPS) and reconstruction (Chamfer distance), as well as each model's capability in single-layer mesh generation, non-watertight reconstruction, and texture mapping extraction. 

As shown in Table \ref{tab:table1} and Fig.~\ref{visual}, our method outperforms state-of-the-art approaches~\cite{huang2DGaussianSplatting2024,chen2024pgsrplanarbasedgaussiansplatting,yuGSDF3DGSMeets2024,yu2024gaussianopacityfieldsefficient} in both rendering quality and geometric accuracy. Specifically for garment reconstruction, it achieves non-watertight reconstruction and single-layer mesh extraction, enabling direct application to fabric simulation and virtual try-on.

\begin{table}
  \centering
  \caption{Quantitative effects of different options. }
  \label{tab:table2}
  \setlength{\tabcolsep}{10mm}
\resizebox{\linewidth}{!}{ 

  \begin{tabular}{l|cccc}
    \toprule
      & SSIM$\uparrow$ & PSNR$\uparrow$ & LPIPS$\downarrow$ & CD$\downarrow$     \\
    \midrule
     Movement &  \cellcolor{yellow}0.965 & 38.52 & \cellcolor{yellow}0.018 & 0.638   \\  
     Rotation &  0.961 & 35.93 & 0.028 & 0.739   \\
     Flattening & \cellcolor{orange}0.964 & 36.45& 0.028 & 0.714    \\
     No-Movement & \cellcolor{yellow}0.965 & 37.23 & \cellcolor{orange}0.027 & 0.723    \\    
     No-Rotation& \cellcolor{red1}0.966 & \cellcolor{yellow}39.47 & \cellcolor{yellow}0.018 & \cellcolor{yellow}0.578  \\
     No-Flattening & \cellcolor{yellow}0.965 & \cellcolor{orange}38.73 & \cellcolor{red1}0.017 & \cellcolor{orange}0.614   \\
     Full & \cellcolor{yellow}0.965 & \cellcolor{red1}40.13 & \cellcolor{red1}0.017 & \cellcolor{red1}0.564   \\
  \bottomrule
    \end{tabular}}
\vspace{-10pt}
\end{table}

\subsection{Ablations}
\label{ssec:subhead}
We conducted an independent evaluation of each component of the Gaussian primitive optimization and designed a series of experiments to measure their effects. Specifically, we evaluated three key aspects of the algorithm: Gaussian movement, rotation, and flattening. Table \ref{tab:table2} shows the quantitative effect of each option.

\section{Conclusion}
\label{sec:conclusion}
We introduce a point cloud-guided Gaussian splatting method for multi-view 3D garment reconstruction. By leveraging dense point clouds to guide the manipulation of Gaussian primitives, our approach enhances the reconstruction of high-fidelity, geometrically accurate garment surfaces while preserving rendering quality. Our evaluation shows that this method surpasses existing state-of-the-art techniques in rendering quality and geometric accuracy, achieving non-watertight reconstruction and enabling single-layer mesh extraction.

\section{Acknowledgments}

This work is supported by the National Natural Science Foundation of China under Grants 61971121.

\bibliographystyle{ACM-Reference-Format}
\balance
\bibliography{sample-base}

\end{document}